\documentclass{article}

\newcommand{\Sys}{Chicle\xspace}

\newcommand{\Fig}[1]{Figure~{\ref{#1}}}
\newcommand{\Tab}[1]{Table~{\ref{#1}}}
\newcommand{\Lst}[1]{Listing~{\ref{#1}}}

\newcommand{\TODO}[1]{} 

\newcommand{\KAU}[1]{} 

\PassOptionsToPackage{numbers}{natbib}
\usepackage[preprint]{nips_2018}

\usepackage[utf8]{inputenc} 
\usepackage[T1]{fontenc}    
\usepackage[hidelinks]{hyperref}       
\usepackage{url}            
\usepackage{booktabs}       
\usepackage{amsfonts}       
\usepackage{nicefrac}       
\usepackage{microtype}      
\usepackage{graphicx}
\usepackage{subfig}
\usepackage{siunitx}
\usepackage{glossaries}
\usepackage{xcolor}
\usepackage{xspace}
\usepackage{listings}
\usepackage{zi4}            

\definecolor{bluekeywords}{rgb}{0.13, 0.13, 1}
\definecolor{greencomments}{rgb}{0, 0.5, 0}
\definecolor{redstrings}{rgb}{0.9, 0, 0}
\definecolor{graynumbers}{rgb}{0.5, 0.5, 0.5}

\usepackage{listings}
\title{Elastic CoCoA: Scaling In to Improve Convergence}

\author{
  Michael Kaufmann$^{\dagger\ddagger}$, Thomas Parnell$^{\dagger}$, Kornilios Kourtis$^{\dagger}$ \\
  $^\dagger$IBM Research, $^\ddagger$Karlsruhe Institute of Technology (KIT) \\
  $^\dagger$Zurich, Switzerland $^\ddagger$Karlsruhe, Germany \\
  \texttt{\{kau,tpa,kou\}@zurich.ibm.com} \\
}

\begin{document}

\maketitle

\begin{abstract}
  In this paper we experimentally analyze the convergence behavior of CoCoA and
  show, that the number of workers required to achieve the highest convergence
  rate at any point in time, changes over the course of the training.  Based on
  this observation, we build \Sys, an elastic framework that dynamically adjusts
  the number of workers based on \emph{feedback from the training algorithm}, in
  order to select the number of workers that results in the highest convergence
  rate.  In our evaluation of 6 datasets, we show that \Sys is able to
  accelerate the time-to-accuracy by a factor of up to $5.96\times$ compared to
  the best static setting, while being robust enough to find an \emph{optimal or
    near-optimal setting automatically} in most cases.
\end{abstract}

\newacronym{cocoa}{CoCoA}{\textbf{Co}mmunication-efficient distributed dual \textbf{Co}ordinate \textbf{A}scent}
\newacronym{sgd}{SGD}{stochastic gradient descent}
\newacronym{sdca}{SDCA}{stochastic dual coordinate ascent}
\newacronym{dg}{DG}{duality-gap}
\newacronym{rdma}{RDMA}{remote direct memory access}
\newacronym{rpc}{RPC}{remote procedure call}
\newacronym{glm}{GLM}{generalized linear model}
\newacronym{ml}{ML}{machine learning}

\section{Introduction}
\label{sec:introduction}

As data has become a major source of insight,
\gls{ml} has become a dominant workload in many (public and private) cloud environments.
Ever-increasing collection of data further drives development of efficient
algorithms and systems for distributed \gls{ml} \cite{smith2018,dunner2018_2} as
resource demands often exceed the capacity of single nodes. However,
distributed execution, and the usage of cloud resources, pose additional challenges in
terms of efficient and flexible resource utilization.  Recently, several works
have aim to improve resource
utilization and flexibility of \gls{ml} applications
\cite{harlap2017,qiao2017,zhang2017}.

In this paper, we focus on \gls{cocoa} \cite{smith2018}, a state-of-the-art
framework for efficient, distributed training of \glspl{glm}.
\gls{cocoa} significantly
outperforms other distributed methods, such as mini-batch versions of \gls{sgd}
and \gls{sdca} by minimizing the amount of communication necessary between
training steps.

Our work is motivated by two characteristics of the \gls{cocoa} algorithm.
First, even assuming perfect scalability and no overheads,
increasing the number of workers $K$ does not, in general, reduce the time to
reach a solution. This is because the convergence rate of \gls{cocoa} degrades
as $K$ increases~\cite{jaggi2014}. Overall, \gls{cocoa} execution is split into
epochs, and increasing $K$ reduces the execution time of each epoch, but also
decreases the \emph{per epoch} convergence rate, requiring more epochs to reach a
solution. Finding the $K$ that minimizes execution time is not trivial and
depends on the dataset.
 
\begin{figure}[h!]
  \centering
  \subfloat[KDDA] {%
    \includegraphics[width=0.30\linewidth]{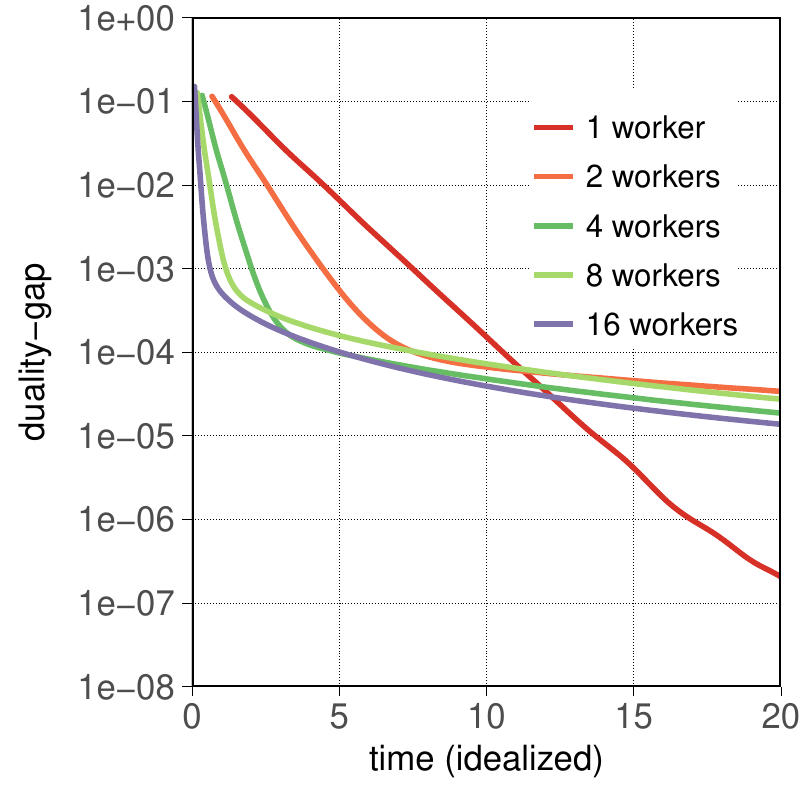}
    \label{fig:intro:kdda}
  }
  \quad
  \subfloat[Higgs] {%
    \includegraphics[width=0.30\linewidth]{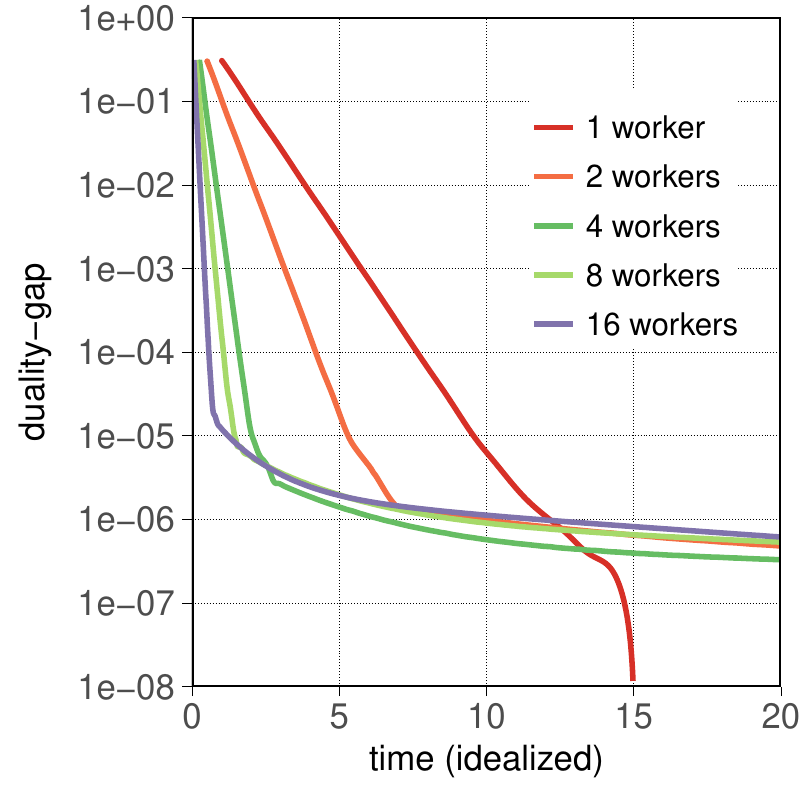}
    \label{fig:intro:higgs}
  }
  \quad
  \subfloat[RCV1] {%
    \includegraphics[width=0.30\linewidth]{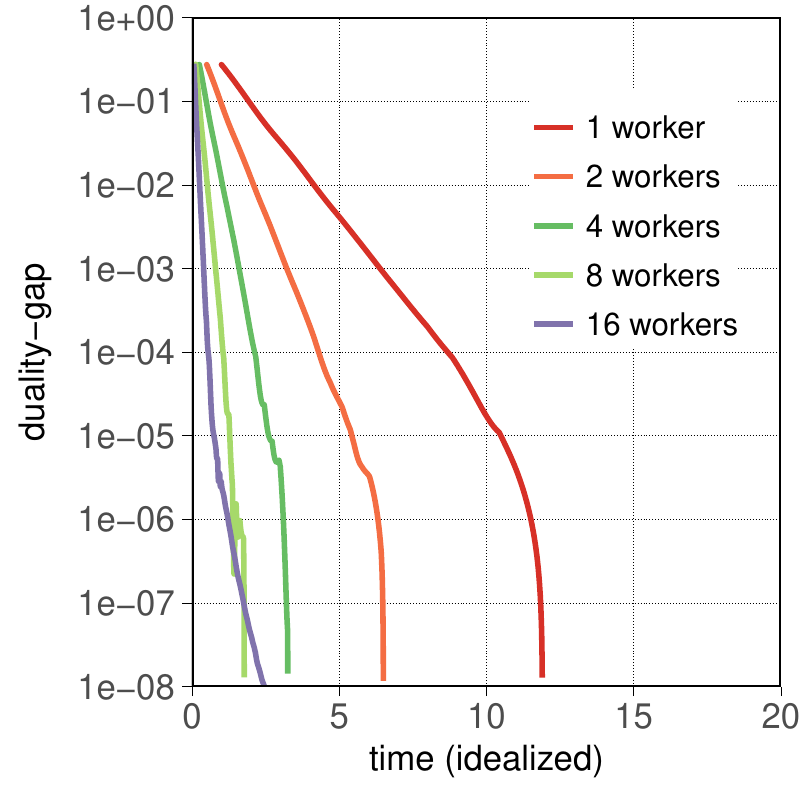}
    \label{fig:intro:rcv1}
  }

  \caption{Example of the convergence of the duality-gap (a certificate for
    accuracy) for 3 datasets using 1 to 16 workers, assuming perfect scaling and
    zero communication cost.}
  \label{fig:intro:idealized}
\end{figure}

Second, the number of workers $K$ that minimize execution time changes as the
algorithm progresses.
\Fig{fig:intro:kdda}/\ref{fig:intro:higgs} shows the convergence rate with
$K=\{1,2,4,8,16\}$ workers, using the kdda and higgs datasets as examples.  We
evaluate the convergence rate by plotting the duality-gap, which is given by the
distance between the primal and dual formulation of the training objective, and
has been shown to provide a robust certificate of convergence \cite{dunner2016,
  smith2018}.
Both examples show that for larger values of $K$, the duality-gap converges
faster initially, but slows down earlier than for smaller values of $K$, thus
resulting in smaller values for $K$ leading to a shorter
time-to-(high)-accuracy\footnote{When we refer to the training accuracy we mean
  that a highly accurate solution to the optimization problem has been found
  (i.e., a small value of the duality gap), rather than the classification
  accuracy of the resulting classifier.}  than large values for $K$. However,
this is not universally true, as \Fig{fig:intro:rcv1} shows for the rcv1
dataset, which scales almost perfectly with $K$.


Based on these observations, we built \Sys, an elastic distributed machine
learning framework, based on \gls{cocoa}, that reduces time time-to-accuracy,
robustly finds (near-)optimal settings automatically and optimizes resource
usage by exploiting the drifting of the optimal $K$.

\section{Background}
\label{sec:background}

\gls{cocoa} \cite{smith2018} is a distributed machine learning framework to
train \glspl{glm} across $K$ workers. The training data matrix $A$ is
partitioned column-wise across all workers and processed by local optimizers
that independently apply updates to a shared vector $v$, which is synchronized
periodically. In contrast to the mini-batch approach, local optimizers apply
intermediate updates directly to their local version of the shared vector $v$,
thus benefiting from previous updates within the same epoch. \KAU{This seems
  very shaky. Tom, can you please have a look at this and/or fix it?}

Due to the immediate local updates to $v$ by local optimizers, \gls{cocoa}
outperforms previous state-of-the-art mini-batch versions of \gls{sgd} and
\gls{sdca}.  However, for the same reason, it is not trivial to efficiently
scale-out \gls{cocoa}, as increasing the number of workers does not guarantee a
decrease in time-to-accuracy, even assuming perfect linear scaling and zero
communication costs between epochs.  The reason for this counter-intuitive
behavior is that, as each local optimizer gets a smaller partition of $A$,
i.e. as it sees a \emph{smaller picture} of the entire problem, the number of
identifiable correlations within each partition decreases as well, thus leaving
more correlations to be identified across partitions, which is slower due to
infrequent synchronization steps.

Moreover, as indicated in the previous section, there is no $K$ for which the
convergence rate is maximal at all times. This poses a challenge about the
selection of the best $K$. It is up to the user to decide in advance whether to
train quickly to a low accuracy and wait longer to reach a high accuracy or vice
versa. A wrong decision can lead to longer training times and wasted resources
as well as money, as resources -- at least in cloud offerings -- are typically
billed by the hour.

Ideally, the system would automatically and dynamically select $K$, such that
the convergence rate is maximal at any point in time, in order minimize training
time and resource waste.  As \Fig{fig:intro:higgs} shows, the convergence rate,
i.e. the slope of the curve, starting from the same level of accuracy, differs
between different settings for $K$. E.g, as the curve for $K=16$ flattens when
reaching $\approx1e-5$, the curves for $K\le8$ become relatively steeper until
they too, one by one, flatten out. Hence, in order to stay within a region of
fast convergence for as long as possible, the system should switch to a smaller
$K$, once the curve for the current $K$ starts to flatten.  We assume that the
convergence rate, when switching from $K$ to $K' < K$ workers, at a certain
level of accuracy, will follow a similar trajectory, as if the training had
reached said level of accuracy starting with $K'$ workers in the first
place. However, the validity of this assumption is obvious, given that the
learned models in both cases are not guaranteed to be indentical.

Apart from the algorithmic side, adjusting $K$ also poses very practical
challenges on the system side. Every change in $K$ incurs a transfer of
potentially several gigabytes of training data between nodes -- a task that
overwhelms many systems \cite{zaharia2010,stuedi2017,sikdar2017} as data
(de-)serialization and transfer can be very time consuming\footnote{Initially,
  we attempted to implement the concept of \Sys in Spark. This, however, failed
  to a large degree due to very time-consuming (de-)serialization of the
  training data.}. It is therefore crucial that the the overhead, introduced by
the adjustment of $K$, is small, such that a net benefit can be realized.

\section{\Sys}
\label{sec:system}

\Sys\footnote{\Sys is the Mexican-Spanish word for latex from the sapodilla tree
  that is used as basis for chewing gum.} is a distributed, auto-elastic machine
learning system based on the state-of-the-art CoCoA \cite{smith2018} framework
that enables efficient \gls{ml} training with minimized time-to-accuracy and
optimized resource usage.
The core concept of \Sys is to reduce the number of workers (and therefore
training data partitions), starting from a set maximum number, dynamically,
based on feedback from the training algorithm.  This is rooted in the
observation of a \emph{knee} in the convergence rate, after which the
convergence slows down significantly, \textbf{and} that this \emph{knee}
typically occurs at a lower duality-gap for fewer workers compared to more
workers. This can be observed in \Fig{fig:intro:higgs}. Here, the \emph{knee}
occurs at $\approx 1e-5$ for 16 workers and $\approx 1e-6$ for 2 workers. The
reasoning for adjusting the number of workers is the assumption that CoCoA can
be accelerated, if, by reducing the number of workers, it can stay before the
\emph{knee} for as long as possible.

\subsection{Overview}
\Sys implements a master/slave design in which a central driver (master)
coordinates one or more workers (slaves), each running on a separate
node. Driver and worker communicate via a custom \gls{rpc} framework based on
\gls{rdma} to enable fast data transfer with minimal overhead. \Sys is
implemented in $\approx$ 3,000 lines of C++ code, including the RDMA-based
\gls{rpc} subsystem.

The \textbf{driver} is responsible for loading, partitioning and distribution
the training data, hence no shared file system is required to store the training
data. It partitions the data into $P \ge K$ partitions for $K$ workers, such
that each worker is assigned $P \over K$ partitions with $P$ being the least
common multiple of $K$ and all potential scale-in sizes $K' < K$. Moreover, the
central \gls{cocoa} component is implemented as driver module.
The \textbf{workers} implement an \gls{sdca} optimizer. Each optimizer instance
works on all partitions assigned to a worker, such that it can train with a
\emph{bigger picture} once partitions get reassigned to a smaller set of
workers. For each epoch, workers compute the partial primal and dual objective
for their assigned partitions, which are sent to the driver where the
duality-gap is computed and passed to the scale-in policy module.

\subsection{Scale-in}
\label{sec:scalein}
\Sys enables efficient adjustment of the number of workers $K$ (and the
corresponding number of data partitions per worker process) using a decision
policy and a \gls{rdma}-based data copy mechanism. In the context of this paper,
\Sys only scales-in, i.e. reduces the number of workers $K$ and therefore
redistributs the number of partitions $P$ across fewer workers.

\begin{figure}
  \centering
  \includegraphics[width=0.30\linewidth]{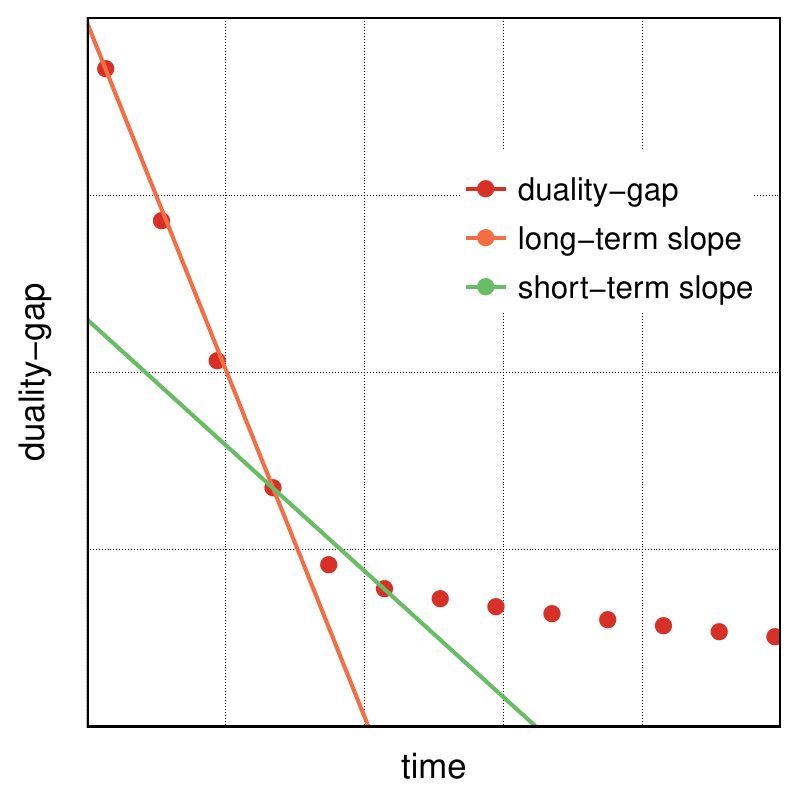}
  \caption{Schematic of the long-/short-term slope of the duality-gap that
    we use to identify the \emph{knee}.}
  \label{fig:sys:policy}
\end{figure}

\paragraph{Scale-in policy.}
\label{sec:sys:policy}
Our scale-in policy attempts to determine the earliest point in time when it is
beneficial to reduce the number of workers $K$ (i.e. the beginning of the
\emph{knee}) while, at the same time, being robust against occasional outlier
(i.e. exceptionally long) epochs.
To that end, we use the slope of the duality-gap over time to identify the
\emph{knee}. It computes two slopes (see \Fig{fig:sys:policy}) -- a long-term
slope $S_l$ which considers the convergence of the duality-gap since the last
scale-in event -- and a short-term slope $S_s$, which considers only the last N
epochs. As soon as $S_s \times d < S_l$ the policy directs the driver process to
initiate the scale-in mechanism. Larger values for $N$ and $d$ generally lead to
a more robust decision w.r.t. occasional outlier epochs, however they also
increase the decision latency, thus potentially failing to maximize benefits
from an earlier scale-in. Empirically, we have determined that $N=2$ and
$d=1.25$ works well across all evaluated datasets.
Our policy does not determine the optimal factor $m$ of the scale-in, i.e.
$K \rightarrow K/m$. We use a fixed $m$ of 4, as test have shown that the
convergence rate difference for smaller $m$ is often very small.

\paragraph{Scale-in mechanism.}
\label{sec:sys:mechanism}
We implement a simple, RDMA-based foreground data-copy mechanism to copy data
from to-be-removed workers to remaining workers. As the data transfer occurs in
parallel, between multiple pairs of workers, we are able to exceed the maximal
single-link bandwidth. For a scale-in from $K$ to $K/m$ workers and a
single-link bandwidth of $r$ (e.g. 10 Gb/s), we can achieve a total transfer
rate of $m \times r$, e.g. 40 Gb/s to scale-in from 16 to 4 workers on a 10 Gb/s
network.

\subsection{Data partitioning and in-memory representation}
While we do not employ a sophisticated data partitioning scheme -- we simply
split the data into equally sized chunks as it is laid out in the input file --
we use an in-memory layout optimized for efficient local access as well as
efficient data transfer between workers (see \Lst{lst:sys:datastructure}). In
\Sys, data for each partition is stored consecutively in the
\texttt{Partition::data} array, which eliminates the need for costly
serialization. On the receiving side, a simple deserialization step is required
to restore the \texttt{Example::dp} pointer into the \texttt{Partition::data}
array for each \texttt{Example}. This data layout, combined with the usage of
RDMA, enables us to transfer data at a rate close to the hardware limit.
\lstset{language=C++, caption={In-memory data structures of \Sys},
  label=lst:sys:datastructure}
\begin{lstlisting}
struct Datapoint { uint32_t feature; float value; };
struct Example   { size_t size; float label; Datapoint *dp; };
struct Partition {
    Example   *examples;    // pointer to examples array in 'data'
    Datapoint *datapoints;  // pointer to datapoints array in 'data'
    double    *model;       // pointer to model vector insize 'data'
    size_t     numExamples; // number of examples
    char      *data;        // contains all data (examples, datapoints, model)
    size_t     size;        // total size of memory allocated for 'data'
};
\end{lstlisting}
While we have considered an anticipatory background transfer mechanism, our
evaluations (see \Tab{tab:eval:overhead}) show that the overhead, introduced by
our mechanism, does not necessitate this.

\section{Evaluation}
\label{sec:evaluation}
In our evaluation, we attempt to answer the question of how much the CoCoA
algorithm can be improved by scaling-in training and thus staying in front of
the \emph{knee} for as long as possible.

\begin{figure}[h!]
  \centering
  \subfloat[RCV1] {%
    \includegraphics[width=0.30\linewidth]{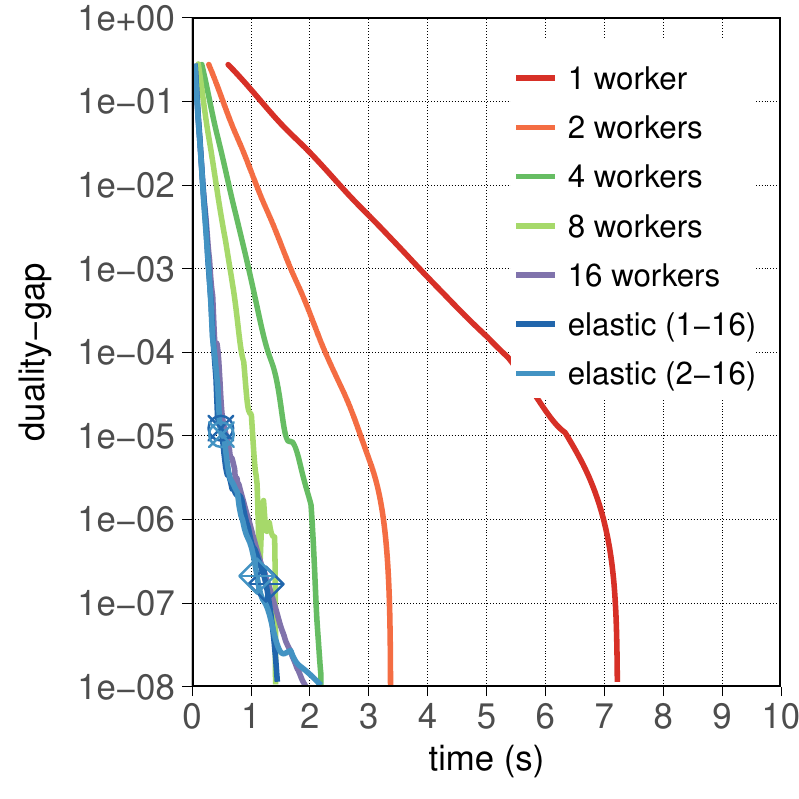}
    \label{fig:eval:rcv1}
  }
  \quad
  \subfloat[KDDA] {%
    \includegraphics[width=0.30\linewidth]{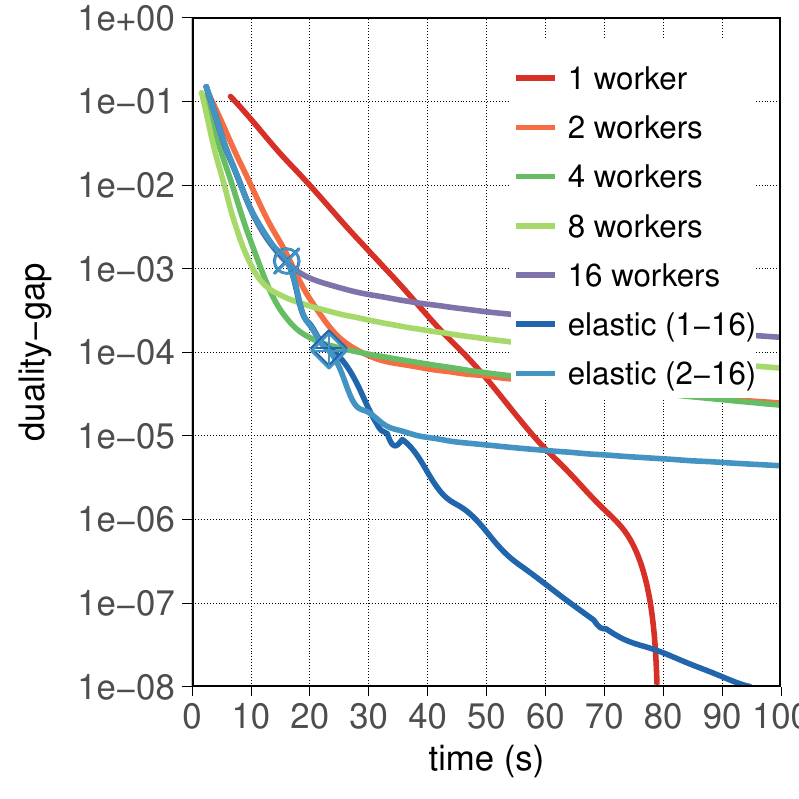}
    \label{fig:eval:kdda}
  }
  \quad
  \subfloat[Higgs] {%
    \includegraphics[width=0.30\linewidth]{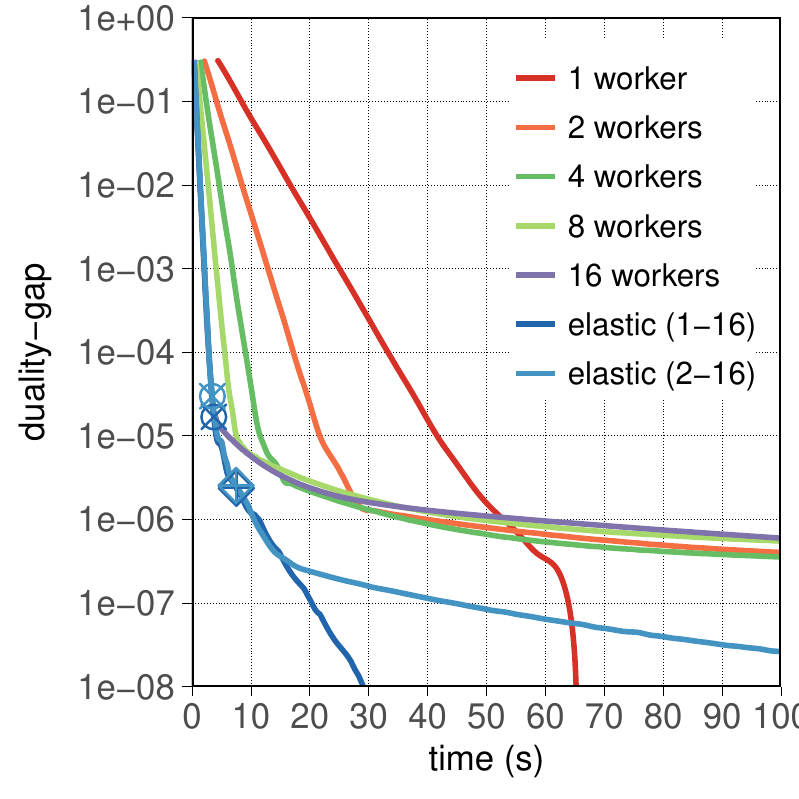}
    \label{fig:eval:higgs}
  }

  \subfloat[KDD12] {%
    \includegraphics[width=0.30\linewidth]{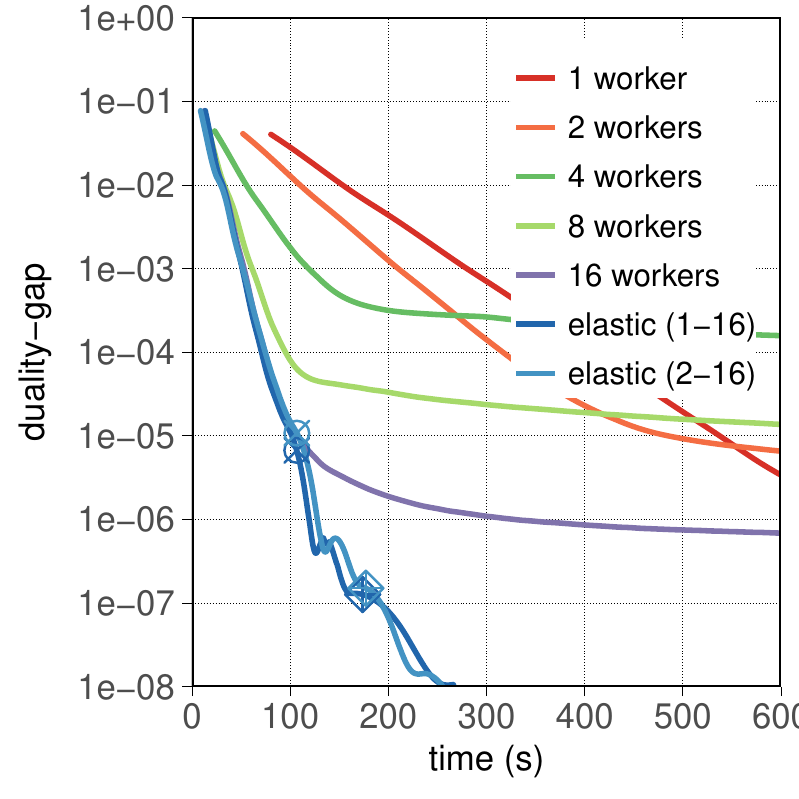}
    \label{fig:eval:kdd12}
  }
  \quad
  \subfloat[Webspam] {%
    \includegraphics[width=0.30\linewidth]{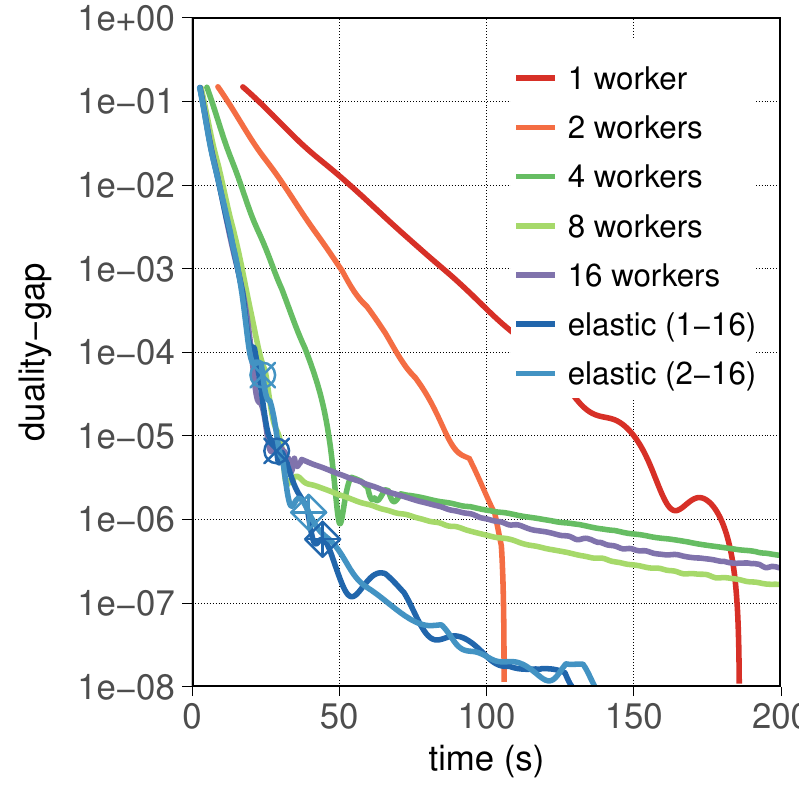}
    \label{fig:eval:webspam}
  }
  \quad
  \subfloat[Criteo] {%
    \includegraphics[width=0.30\linewidth]{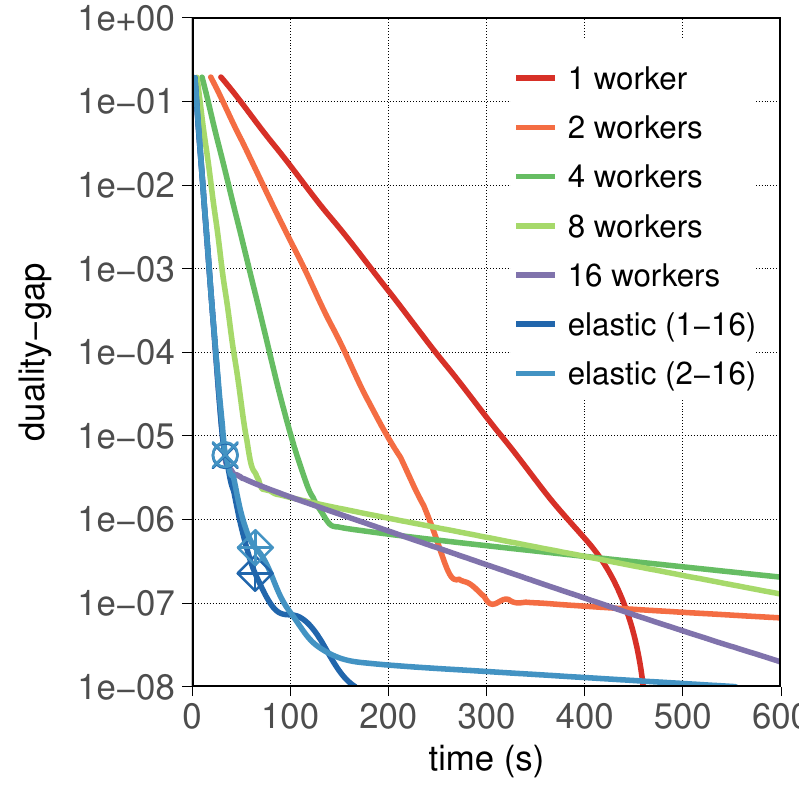}
    \label{fig:eval:criteo}
  }
  \caption{Duality-gap vs. time plots for the evaluated datasets and
    settings. Circles depict a scale-in from 16 to 4 workers, diamonds a
    scale-in from 4 to 2 and 1 worker(s), respectively.}
  \label{fig:eval}
\end{figure}

To answer this question, we compare the time-to-accuracy (duality-gap) of our
static CoCoA implementation with our elastic version, using an SVM training
algorithm\footnote{we use a constant regularizer term $\lambda=0.01$} and the 6
datasets shown in \Tab{tab:eval:datasets}. We evaluate static settings with 1,
2, 4, 8 and 16 workers as well as two elastic settings. In the first elastic
setting, we start with 16 workers and scale-in to a single worker. This
represents cases where the entire dataset fits inside a single node's memory but
limited CPU resources make distribution beneficial anyway. In the second elastic
setting, we start with 16 workers but scale-in to only two workers. This
represents cases where a dataset exceeds a single node's memory capacity and
therefore has to be distributed. As convergence behavior for 2+ nodes is similar
(see \Fig{fig:eval}), this also indicates how our method works in a larger
cluster, e.g., when scaling from 64 to 8 nodes.
\begin{table}[!h]
  \centering
  \begin{tabular}{l|rrrr}
    \toprule
    Dataset &   Examples &    Features & Size (SVM) & Sparsity   \\
    \midrule
    RCV1    &    667,399 &      47,236 & 1.2 GB     &    0.16 \% \\
    KDDA    & 20,216,830 &   8,407,752 & 2.5 GB     & 1.8e-04 \% \\
    Higgs   & 11,000,000 &          28 & 7.5 GB     &   92.11 \% \\
    KDD12   & 54,686,452 & 149,639,105 &  21 GB     &   2e-05 \% \\
    Webspam &    350,000 &  16,609,143 &  24 GB     &    0.02 \% \\
    Criteo  & 45,840,617 &     999,999 &  35 GB     & 3.9e-03 \% \\
    \bottomrule
  \end{tabular}
  \caption{Datasets used in the evaluation}
  \label{tab:eval:datasets}
\end{table}
All test are executed on a 17 node cluster, equipped with Intel Xeon
E5-2640v3/E5-2650v2, 160-256 GB RAM and CentOS/Fedora 26 Linux, running 16
workers and 1 driver, connected by a FDR (56 Gb/s) Infiniband fabric.  The
initial set of nodes is always chosen randomly.  The results, shown in
\Fig{fig:eval} and \Tab{tab:eval:speedup}, represent the best results over 6
test runs for all schemes, to account for potential node speed variations. We
set a test time limit of 10 minutes (not including data loading).
Time results include computing the duality gap.

Our evaluation shows that the basic concept of \Sys ~-- to adjust the number of
workers based on feedback from the training algorithm -- has benefits for most
evaluated datasets.
When scaling down to a single worker, \Sys shows an average speedup of 2$\times$
compared to the best static setting and 2.2$\times$ when scaling down to two
workers. While our method does not improve upon all evaluated settings and
target accuracies (e.g., $1e-8$ for KDDA, Webspam, RCV1), the slowdown (compared
to the respective best static setting) is tolerable, and speedups are still
achieved compared to non-optimal static settings. It is important to note that
the optimal static setting is not necessarily known in advance and may require
several test runs to determine. \Sys, on the other hand, is able to find an
optimal or near optimal setting automatically, which shows its robustness.

\begin{table}
  \subfloat[1-16 workers] {%
  \label{tab:eval:speedup1}
  \centering
  \begin{tabular}{l|rrr}
    \toprule
    \cmidrule(r){1-4}
    Dataset & $1e-6$   & $1e-7$   & $1e-8$      \\
    \midrule
    RCV1    & 1.05 (16) & 1.06 (16) & 0.98  (8) \\
    KDDA    & 1.49  (1) & 1.12  (1) & 0.83  (1) \\
    Higgs   & 3.21  (4) & 3.14  (1) & 2.24  (1) \\
    KDD12   & 2.75 (16) & >3.15     & >2.25     \\
    Webspam & 1.25  (4) & 1.43  (2) & 0.82  (2) \\
    Criteo  & 2.82  (4) & 3.80  (2) & 2.76  (1) \\
    \bottomrule
  \end{tabular}
}
  \quad
  \subfloat[2-16 workers] {%
  \label{tab:eval:speedup2}
  \centering
  \begin{tabular}{l|rrr}
    \toprule
    \cmidrule(r){1-4}
    Dataset & $1e-6$   & $1e-7$   & $1e-8$      \\
    \midrule
    RCV1    & 1.31 (16) & 1.12 (16) & 0.64  (8) \\
    KDDA    & >1.28     & --        & --        \\
    Higgs   & 3.46  (4) & 5.96 (16) & >3.63     \\
    KDD12   & 2.57 (16) & >3.12     & >2.35     \\
    Webspam & 1.12  (4) & 1.59  (2) & 0.77 (2)  \\
    Criteo  & 2.63  (4) & 3.23  (2) & >1.08     \\
    \bottomrule
  \end{tabular}
  }
  \caption{Speed-up factor of an elastic vs. the best static setting (the number
    of workers of the best static setting is given in parentheses) for reaching
    a target accuracy of $1-e6$ -- $1e-8$. In case no static setting has reached
    the target accuracy within a 10 minute time-limit, we provide a minimum
    speedup factor and ``--'' in case neither an elastic, nor a static setting
    has reached a target accuracy.}
  \label{tab:eval:speedup}
\end{table}

\begin{table}
  \centering
  \begin{tabular}{l|cccccc}
    \toprule
    \cmidrule(r){1-6}
    Setting       & RCV1    & KDDA    & Higgs   & KDD12   & Webspam & Criteo   \\
    \midrule
    1-16 workers  & 0.12 s  & 0.73 s  & 0.71 s  & 5.04 s  & 2.78 s  & 4.52 s   \\
    2-16 workers  & 0.06 s  & 0.39 s  & 0.38 s  & 2.78 s  & 1.53 s  & 2.18 s   \\
    \bottomrule
  \end{tabular}
  \caption{Total average scale-in overhead}
  \label{tab:eval:overhead}
\end{table}

Finally, we measured data-copy rates and overhead due to scaling-in. Both
metrics include the actual data-transfer, control plane overhead and data
deserialization. We measured data transfer rates of up to 5.8 GiB/s (1.4 GiB/s
on average) and overheads as shown in \Tab{tab:eval:overhead}.  As the measured
times do not constitute a significant overhead on our system, we did not
implement background data transfer. For slower networks, such a method could be
used to hide data transfer times behind regular computation.

\KAU{I would like to say that this method is able to accelerate datasets that
  are strongly correlated, which are the ones that I expect are typically
  difficult to parallelize. For ones that don't have that many strong
  correlations, the current CoCoA works fine. Is that a valid statement? Problem
  is, I don't know how strong or weak the correlations in each dataset are.}



\section{Related Work}
\label{sec:related_work}
To our knowledge \Sys is the first elastic \gls{cocoa} implementation.
Several other elastic \gls{ml} systems exist, but
in contrast to \Sys,
they target efficient resource utilization rather than reducing overall execution
time.
Litz \cite{qiao2017} is an elastic ML framework that over-partitions training
data into $P = n \times K$ partitions for $K$ physical workers. Elasticity is
achieved by increasing or decreasing the amount of partitions per node. In
contrast to \Sys, Litz does not scale based on feedback from the training
algorithm nor does it improve the per-epoch training algorithm convergence rate
when doing so, as partitions are always processed independently of each other.
SLAQ \cite{zhang2017} is a cluster scheduler for \gls{ml} applications. Like
\Sys, SLAQ uses feedback from \gls{ml} applications, but instead of optimizing
the time to arbitrary accuracy for one application, SLAQ tries minimize the time
to low accuracy for many applications at the same time, by shifting resources
from applications with low convergence ratse to those with high ones, assuming
that resources can be used more effectively there.
Proteus \cite{harlap2017} enables the execution of ML applications using
transient revocable resources, such as EC2's spot instances, by keeping worker
state minimal at the cost of increased communication.

\section{Conclusion and Future Work}
In this paper we have shown experimentally, that the optimal number of workers
for \gls{cocoa} changes over the course of the training. Based on this
observation we built \Sys, an elastic ML framework, and have shown that it can
outperform static \gls{cocoa} for several datasets and settings by a factor of
2--2.2$\times$ on average, often, while using fewer resources. Future work
includes additional ways to dynamically optimize
\gls{cocoa} in terms of training time and resource usage, as well as related
use-cases, e.g. neural networks \cite{lin2018}. Furthermore, we are working
towards a theoretical foundation of our observations.

\newpage
{\small
  \bibliographystyle{acm}
  \bibliography{bibliography.bib}
  \label{sec:references}
}
%
%
%
%
%
%
\end{document}